\def\year{2019}\relax
\documentclass[letterpaper]{article} 
\usepackage{aaai19}  
\usepackage{times}  
\usepackage{helvet}  
\usepackage{courier}  
\usepackage{url}  
\usepackage{graphicx}  

\usepackage{multirow}
\usepackage{amsmath}
\usepackage{bm}

\begin{document}
%
\title{Promoting Diversity for End-to-End Conversation Response Generation}

\author{Yu-Ping Ruan$^1$, Zhen-Hua Ling$^1$, Quan Liu$^{1,2}$, Jia-Chen Gu$^1$, Xiaodan Zhu$^3$\\
$^1$National Engineering Laboratory for Speech and Language Information Processing,\\
University of Science and Technology of China, Hefei, P.R.China\\
$^2$iFLYTEK Research, Hefei, P.R. China\\
$^3$Department of Electrical and Computer Engineering, Queen's University, Kingston, Canada\\
\tt \small \{ypruan,gujc\}@mail.ustc.edu.cn, \{zhling,quanliu\}@ustc.edu.cn, xiaodan.zhu@queensu.ca}

\maketitle
\begin{abstract}
We present our work on Track 2 in the Dialog System Technology Challenges 7 (DSTC7). The DSTC7-Track 2 aims to evaluate the response generation of fully data-driven conversation models in knowledge-grounded settings, which provides the contextual-relevant factual texts. The Sequence-to-Sequence models have been widely used for end-to-end generative conversation modelling and achieved impressive results. However, they tend to output dull and repeated responses in previous studies. Our work aims to promote the diversity for end-to-end conversation response generation, which follows a two-stage pipeline: 1) Generate multiple responses. At this stage, two different models are proposed, i.e., a variational generative ({VariGen}) model and a retrieval based ({Retrieval}) model. 2) Rank and return the most related response by training a topic coherence discrimination (TCD) model for the ranking process.
According to the official evaluation results, our proposed Retrieval and VariGen systems ranked first and second respectively on objective diversity metrics, i.e., Entropy, among all participant systems. And the VariGen system ranked second on {NIST} and {METEOR} metrics.
\end{abstract}

\begin{figure*}[!t]
	\centering
	\includegraphics[width=6.2in]{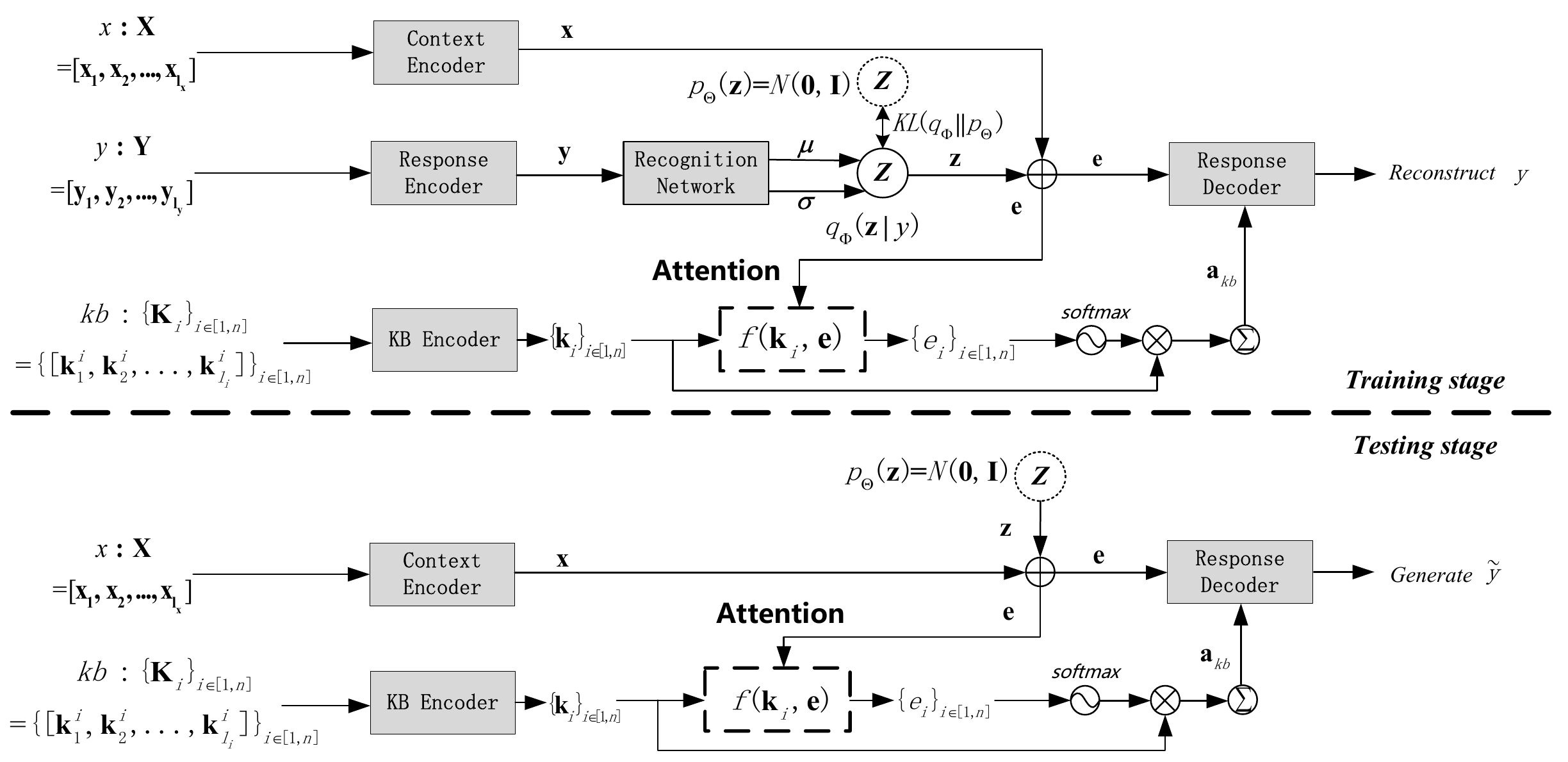}
	\caption{The model architecture of the variational generative model (VariGen) implemented in this paper. $\bigoplus$ denotes the concatenation of input vectors. All the encoders and decoders are 1-layer GRU-RNNs, the recognition network is a two-layer MLP.}
	\label{fig:models}
\end{figure*}

\section{Introduction}\label{intro}

 Natural language conversation occupies one of the most challenging tasks in artificial intelligence (AI) and natural language processing (NLP), which can be categorized into task oriented dialog systems \cite{young2013pomdp} and non-task oriented chatbots. Previous studies in building dialog systems mainly focused on either rule-based or learning-based methods \cite{DBLP:journals/ker/SchatzmannWSY06,young2013pomdp,DBLP:journals/corr/BordesW16,DBLP:journals/corr/WenGMRSUVY16}. These methods often require manual efforts in rule designing and feature engineering, which makes it difficult to develop an extensible open domain conversation system.

 Recently due to the explosive growth of social media, there comes vast amount of conversation text available on the web. This makes the data-driven approach to settle the conversation problem possible. Thus, there has been a growing interest in applying encoder-decoder models \cite{sutskever2014sequence,DBLP:journals/corr/VinyalsL15,shang2015neural,serban2016building} for conversation in a completely end-to-end data-driven fashion, which have produced impressive results.

 The Dialog System Technology Challenges (DSTC) in its seventh edition offers a track (Track 2) \cite{dstc7_track2} devoted exclusively to fully data-driven approaches for conversation modelling. Different from previous data-driven dialog systems which mostly focused on chitchat, DSTC7-Track 2 tries to push the data-driven conversation models beyond chitchat in order to produce system responses that are both substantive and ``useful'' which can contain factual contents. So DSTC7-Track 2 provides not only the social conversation corpus but also contextual-related factual texts to build a knowledge-grounded conversation settings.

 Among those neural end-to-end models for data-driven conversation modelling, the encoder-decoder framework has been widely adopted and they principally learn the mapping from an input conversation context $x$ to its target response $y$.
 However, previous studies on generating responses for chit-chat conversations \cite{serban2016building,li2015diversity}
have found that ordinary encoder-decoder models tend to generate dull, repeated and generic responses in conversations,  such as \emph{``i don't know", ``that's ok"}, which are lack of diversity.
One possible reason is the deterministic calculation of ordinary encoder-decoder models which constrains them
from learning the $1$-to-$n$ mapping relationship, especially on semantic connections, between input sequence and potential multiple target sequences, which is common in social media conversation corpus.

To cope with the problem of lack diversity, we built our systems following a two-stage processing pipeline: 1) The first module (module-1) outputs multiple candidate responses according to input context and related textual facts. 2) The second module (module-2) ranks the output responses from module-1 and returns the most topic relevant response. Two different models are designed for module-1, i.e., one Variational generative model and one retrieval based model. Variational generative models are suitable for learning the $1$-to-$n$ mapping relationship due to their variational sampling mechanism for deriving latent representations, and recently they have been applied to dialog response generation \cite{zhao2017learning,serban2017hierarchical}. For module-2, a topic coherence discrimination (TCD) model is designed and trained based on ESIM \cite{chen2017enhanced} for the ranking process.
According to the official evaluation results, our proposed Retrieval and VariGen system ranked first and second respectively on objective diversity metrics, i.e., Entropy, among all participant systems, and the VariGen system ranked second on {NIST} and {METEOR} metrics.

\section{System Description}\label{model}
As we have described in Section \ref{intro}, we designed our systems following a two-stage processing pipeline: 1) The first module (module-1) outputs multiple possible responses according to input context and related textual facts. 2) The second module (module-2) ranks the output responses from module-1 and return the most topic relevant response. Here, we will describe our models in detail.

\subsection{Candidate response generation}
At this stage, a generative model based on Variational AutoEncoder (VAE) and a retrieval model based on Bag-of-Words (BoW) representation are designed as module-1 in order to output multiple responses with diversity.
\paragraph{Variational generative model (VariGen)}
The model architecture of the variational generative model (VariGen) implemented in this paper is shown in Figure \ref{fig:models}. For an input conversation context $x=[x_1, x_2, ...,x_{l_x}]$ with $l_x$ words, we can derive the corresponding output hidden states $[\mathbf{h}_1, \mathbf{h}_2, ...,\mathbf{h}_{l_x}]$ by sending its word embedding sequence $\mathbf{X}=[\mathbf{x}_1, \mathbf{x}_2, ...,\mathbf{x}_{l_x}]$ into the \emph{Context Encoder}.
Then, the mean pooling of hidden states $[\mathbf{h}_1, \mathbf{h}_2, ...,\mathbf{h}_{l_x}]$ is used to present the input context, denoted as $\mathbf{x}$.
Similarly, for its correlated textual facts $kb=[k_1, k_2, ..., k_n]$ with $n$ textual facts in it, and each $k_i$ in $kb$ is also a word sequence, i.e., $k_i=[k_{1}^i, k_{2}^i, ...,k_{l_{i}}^i]$ with $l_{i}$ words, we can derive vector representation $\mathbf{k_i}$ for each $k_i$ by inputting its word embedding sequence $\mathbf{K}_i=[\mathbf{k}_1^i, \mathbf{k}_2^i, ...,\mathbf{k}_{l_i}^i]$ into the \emph{KB Encoder}. Also, we can derive vector representation $\mathbf{y}$ for response $y$ by inputting $\mathbf{Y}=[\mathbf{y}_1, \mathbf{y}_2, ...,\mathbf{y}_{l_y}]$ into the \emph{Response Encoder}.

The \emph{Recognition Network} is a multi-layer perceptron (MLP), which has a hidden layer with \emph{softplus} activation and a linear output layer in our implementation.
The recognition network predicts $\mathbf{\bm{\mu}}$ and $\log(\mathbf{\bm{\sigma}}^2)$ from $\mathbf{y}$, which gives $q_\phi(\mathbf{z}|y)=\mathcal{N}\mathbf{(\bm{\mu}, {\bm{\sigma}}^2I)}$.
Then samples of latent variable $\mathbf{z}$ are generated from the $q_\phi(\mathbf{z}|y)$ at training stage or directly from $\mathcal{N}\mathbf{(0, I)}$ at testing stage.
To guarantee the feasibility of error backpropagation for model training,
reparametrization \cite{kingma2013auto} is performed to generate the samples of $\mathbf{z}$.

Then we derive the encoding vector $\mathbf{e}$ by concatenating the $\mathbf{z}$ and $\mathbf{x}$, which combines the information of conversation context $x$ and response $y$. And we use $\mathbf{e}$ to extract the key information in correlated $kb$ by using attention on $[\mathbf{k}_1, \mathbf{k}_2, ..., \mathbf{k}_n]$ as follows,

\begin{align}
 e_i &= \mathbf{v}^T \tanh(\mathbf{W}[\mathbf{e};\mathbf{k}_i]),\\
 \mathbf{a}_{kb} &= \sum_{i=1}^{n} \frac{\exp{e_{i}}}{\sum_{j=1}^{n} \exp{e_{j}}} \mathbf{k}_i,
\end{align}
where $\mathbf{a}_{kb}$ is the final attentive output vector from input $kb$.
Finally, for the $Response Decoder$, its initial hidden state is $\mathbf{x}$.
In each time step, its input is the concatenation of the word embedding from the previous time step, the encoding vector $\mathbf{e}$, and the attentive $kb$ vector $\mathbf{a}_{kb}$.

Our VariGen model can be efficiently trained with the stochastic gradient variational Bayes (SGVB) \cite{kingma2013auto} framework by maximizing the lower bound of the conditional log likelihood $\log p(y|x, kb)$ as follows,

\begin{equation}\label{eq:loss}
\begin{aligned}
\mathcal{L}(\theta,\phi;x,y, kb)&=-KL(q_\phi(\mathbf{z}|y)||\mathcal{N}\mathbf{(0, I)})\\
&+\mathbf{E}_{q_\phi(\mathbf{z}|y)}[\log p_\theta(y|x,\mathbf{z},kb)]\\
&\le \log p(y|x,kb).
\end{aligned}
\end{equation}

 in which the summation of the log-likelihood of reconstructing $y$ from the $Response Decoder$ and the negative KL divergence between $q_\phi(\mathbf{z}|y)$ and prior distribution $\mathcal{N}(\mathbf{0}, \mathbf{I})$ is used as the objective function for training.

 When generating responses at testing stage, $k$ samples of $\mathbf{z}$ are generated from $\mathcal{N}(\mathbf{0}, \mathbf{I})$. Then, for each $\mathbf{z}$ sample, a beam-search is adopted to return the top-1 result. The final $k$ generated results will be input to TCD model for subsequent ranking.

\paragraph{Retrieval model}
The retrieval model we built in this work is mainly based on Bag-of-Word (BoW) representation. Specifically, for each input conversation context $x=[x_1, x_2, ...,x_{l_x}]$ with $l_x$ words, we derive the vector representation $\mathbf{x}$ as a weighted sum of its word embedding sequence $\mathbf{X}=[\mathbf{x}_1, \mathbf{x}_2, ...,\mathbf{x}_{l_x}]$, in which each word's inverse-document-frequency (idf) is used as its sum weight and the GloVe embeddings \cite{pennington2014glove} are used. To cope with variable length of $x$, we normalize the final weighted BoW vector $\mathbf{x}$ with the sum of all corresponding idf weights. Then we treat input conversation context $x_{query}$ as a query and retrieve related conversation context $x_{pool}$ in training pool according to cosine similarity between weighted BoW representation $\mathbf{x}_{query}$ and $\mathbf{x}_{pool}$. We do not use the $kb$ in our retrieval model. Finally, we return corresponding responses of top-$k$ related $x_{pool}$ as outputs.

\subsection{Response reranking}

The multiple candidate responses output from module-1 are reranked using a topic coherence discrimination (TCD) model, which is designed and trained based on the ESIM model \cite{chen2017enhanced}.
Specifically, we replace all BiLSTMs in the ESIM with 1-layer LSTMs and define the objective of the TCD model as judging whether a response is a valid response to a given conversation context.
In order to train the TCD model, all context-response pairs in the training set are used as positive samples and negative samples are constructed by randomly shuffling the mapping between contexts and responses. Finally, ranking scores are adopted to rerank all responses generated for one context.
The scores are calculated as: 
\begin{equation} \label{eq:score}
score = \log p_\theta(\tilde{y}|x,kb) + \lambda * \log{p_{TCD}(true|x,\tilde{y})},
\end{equation}
where the first term is the log-likelihood of generating response $\widetilde{y}$ using the decoder network in the VariGen models (when module-1 is the retrieval model, this term will be omitted). The second term is the log-likelihood of the output probability of the TCD model.
$\lambda$ represents the weight between  these two terms.

\section{Experiments}
\subsection{Datasets}
The training, development, and final test datasets of DSTC7-Track 2 were built using the official scripts\footnote{\url{https://github.com/DSTC-MSR-NLP/DSTC7-End-to-End-Conversation-Modeling}}, we downloaded the formatted Reddit conversation data and its corresponding textual facts. We finally collected $1,918,146$, $108,600$, and $10,808$ samples for training, development and final test respectively.

To alleviate the problem of out-of-vocabulary (oov), we replaced all \emph{date, time, email, numbers, link} and other data words with special tokens like $\langle data\rangle, \langle time\rangle$ and so on using \emph{CommonRegex}\footnote{\url{https://github.com/madisonmay/CommonRegex}} tool. We lowercased all words and filtered the \emph{kb} with no more than 5 words. Finally, the corresponding $kb$ of each \emph{context-response} pair has averagely $152$ items in it and each kb item has averagely $33$ words. We set the vocabulary size as $51,996$ and all the oov rates of \emph{contexts}, \emph{responses}, and \emph{kb} in training dataset are no more than $3\%$.

\subsection{Parameter setting}
We trained the VariGen model in our experiments with the following hyperparameters.
All word embeddings, hidden layers of the recognition network and hidden state vectors of the encoders and decoders had $300$ dimensions. The latent variables $\mathbf{z}$ had $100$ dimensions.
All encoder and decoder in VariGen shared one set of embeddings, and the vocabulary size was $51,996$. All model parameters were initialized randomly with Gaussian-distributed samples except for word embeddings were initialized with Glove embeddings \cite{pennington2014glove}. The method of Adam\cite{kingma2014adam} was adopted for optimization. The initial learning rate was $1e-04$. Gradient clipping was set to $1$, and the batch size was $16$.
We generated multiple responses for each post, the number of $\mathbf{z}$ samples was set to 50, the beam search size was 5.
 When training the TCD model, we adopted the same parameter settings above for training VariGen model, and the weight $\lambda$ for reranking was heuristically set to $10$.

\subsection{Baseline models}
We listed three baseline systems provided by the official organizer of DSTC7-Track2 and their evaluation results will be used to help us analyse the performance of our proposed systems.
\begin{itemize}
\item \emph{Constant:} This system always return ``i don't know what you mean.'' as responses.
\item \emph{Random:} This system always randomly picks up a response from training dataset.
\item \emph{Seq2Seq:} This is a GRU-based sequence-to-sequence (Seq2Seq) generative system. The model does not use grounding information ("facts"), attention or beam search. It uses greedy decoding (unkown token disabled).
\end{itemize}

\begin{table*}[!t]

	\centering
	\begin{tabular}{l|cccc|cccc|c}
		\hline
        \hline
		\bf{Models} & nist1 & nist2 & nist3 & nist4 & bleu1 & bleu2 & bleu3 & bleu4 & Meteor \\
        \hline
		Constant &0.175	&0.183	&0.184	&0.184	&39.7\%	&12.8\%	&6.06\%	&2.87\%	&7.48\% \\
        Random & 1.573	&1.633	&1.637	&1.637	&26.4\%	&6.7\%	&2.24\%	&0.86\%	&5.91\% \\
        Seq2Seq & 0.849	&0.910	&0.915	&0.916	&45.2\%	&14.8\%	&5.23\%	&1.82\%	&6.96\% \\
        \hline
        \emph{Retrieval} & 1.938 &2.034	&2.039	&2.040	&29.2\%	&8.2\%	&2.81\%	&1.05\%	&7.48\% \\
        \emph{VariGen} & 2.181	&2.312	&2.322	&2.322	&34.9\%	&10.6\%	&3.67\%	&1.21\%	&7.18\% \\
		\hline
        {Human} & 2.424 &2.624	&2.647	&2.650	&34.1\%	&12.4\%	&5.72\%	&3.13\%	&8.31\% \\
        \emph{Best system performance$^\ast$} & 2.341	&2.510	&2.522	&2.523	&41.3\%	&14.4\%	&5.01\%	&1.94\%	&8.07\% \\
        \hline
		\hline
	\end{tabular}
    \caption{The official \emph{objective evaluation} results of three baseline systems, our two proposed systems, and human results on 2,208 samples from test set. The \emph{Best system performance$^\ast$} represent the best results on each metric among all participant systems.}
	\label{tab:object}
\end{table*}

\begin{table*}[!t]

	\centering
	\begin{tabular}{l|cccc|cc|c}
		\hline
        \hline
		\bf{Models} & entropy1 & entropy2 & entropy3 & entropy4 & div1 & div2 & avg\_len\\
        \hline
		Constant &2.079	&1.946	&1.792	&1.609	&0.000	&0.000	&8.000 \\
        Random & 6.493	&9.670	&10.403	&10.467	&0.160	&0.647	&19.192 \\
        Seq2Seq & 3.783	&5.017	&5.595	&5.962	&0.014	&0.048	&10.604 \\
        \hline
        \emph{Retrieval} & 6.360	&9.374	&10.009	&10.057	&0.108	&0.449	&22.336 \\
        \emph{VariGen} & 5.320	&8.080	&9.487	&10.016	&0.034	&0.265	&16.570 \\
		\hline
        {Human} & 6.589	&9.742	&10.410	&10.445	&0.167	&0.670	&18.757 \\
        \emph{Best system performance$^\ast$} & 6.360	&9.374	&10.009	&10.057	&0.121	&0.449 & --- \\
        \hline
		\hline
	\end{tabular}
    \caption{The official \emph{objective diversity evaluation} results of three baseline systems, our two proposed systems, and human results on 2,208 samples from test set. The \emph{Best system performance$^\ast$} represent the best results on each metric among all participant systems.}
	\label{tab:object_diversity}
\end{table*}

\begin{table*}[!t]
	\centering
	\begin{tabular}{l|cc|cc|cc}
		\hline
        \hline
		\multirow{2}{*}{\bf{Models}} & \multicolumn{2}{c|}{\bf{(A) Relevance}} & \multicolumn{2}{c|}{\bf{(B) Interest}} & \multicolumn{2}{c}{\bf{(C) Overall}}\\
		\cline{2-7}
		& Mean Score & 95\% CI & Mean Score & 95\% CI & Mean Score & 95\% CI \\
		\hline
        Constant & 2.60	&(2.560, 2.644)	&2.32	&(2.281, 2.364)	&2.46	&(2.424, 2.500) \\
        Random & 2.32	&(2.269, 2.371)	&2.35	&(2.303, 2.401)	&2.34	&(2.288, 2.384)\\
        Seq2Seq & 2.91	&(2.858, 2.963)	&2.68	&(2.632, 2.730)	&2.80	&(2.748, 2.844) \\
        \hline
        \emph{Retrieval} & 2.82	&(2.771, 2.870)	&2.57 &(2.525, 2.619)	&2.70	&(2.650, 2.742 )\\
        \emph{VariGen} &--- & ---& ---& ---& ---& ---\\
        \hline
        Human & 3.61 &(3.554, 3.658)	&3.49	&(3.434, 3.539)	&3.55	&(3.497, 3.596)\\
		\hline
        \hline
	\end{tabular}
    \caption{The official \emph{subjective evaluation} results of three baseline systems, our two proposed systems, and human results on 1,000 samples from test set.}
	\label{tab:subject}
\end{table*}

\subsection{Results and analysis}
\paragraph{Objective evaluation}
As Table \ref{tab:object} shows, we present official evaluation results on {NIST, BLEU,} and {METEOR} metrics of three baseline systems, our \emph{Retrieval} and \emph{VariGen} systems, human results, and best system performance among all participants. These objective evaluations were performed on 2,208 samples from final test dataset. For {BLEU} metrics, we can find \emph{Constant} system performed best on the whole among all systems though this system always output ``i don't know what you mean.'', which indicates the limitations of {BLEU} metrics. For {METEOR} metric, we can find that both our proposed \emph{VariGen} and \emph{Retrieval} systems outperformed \emph{Seq2Seq} and \emph{Random} systems. However, \emph{Constant} system also obtained best performance among all systems except for \emph{Human} and {best system performance}$^\ast$, which also indicates the limitations of {METEOR} metric. {NIST}\footnote{\url{ftp://jaguar.ncsl.nist.gov/mt/resources/mteval-v14c.pl}} is an variant of {BLEU} which instead of giving equal weight to each n-gram match but calculates how informative each particular n-gram is \cite{doddington2002automatic}.
We can find that our \emph{Retrieval} and \emph{VariGen} systems outperformed all three baseline systems, and \emph{VariGen} performed better than \emph{Retrieval} system. Also, we can find that \emph{Random} system outperformed \emph{Seq2Seq}, which may be attribute to that the \emph{Seq2Seq} system always produces blank and non-sense responses with very low informativeness. According to the official evaluation results, our proposed \emph{VariGen} system ranked second on automatic evaluation metrics {NIST} and {METEOR} among all participant systems.

As for the objective diversity evaluation, we present official evaluation results as Table \ref{tab:object_diversity}
shows. The div1 and div2, a crude diversity measure of distinct 1-gram and 2-gram respectively as explained in \cite{li2015diversity}, are also known as distinct-1 and distinct-2. Entropy is believed to be a more principled measure of diversity as explained in \cite{zhang2018generating}. We can found that our \emph{Retrieval} and \emph{VariGen} outperformed all baseline models on div1/2 and Entropy metrics except for the \emph{Random} system. It's reasonably that \emph{Random} system has a good performance on diversity metrics since it always randomly select a human response from training dataset. Our \emph{Retrieval} system outperformed \emph{VariGen}, but in Table \ref{tab:object} \emph{VariGen} performed better than \emph{Retrieval} on NIST and BLEU metrics.  According to the official evaluation results,
our \emph{Retrieval} system ranked first on diversity metrics and \emph{VariGen} system ranked second on Entropy metrics.

\paragraph{Subjective evaluations}
Table \ref{tab:subject} presents the official subjective evaluation results which were performed on 1,000 samples from final test dataset. According to official organizer's description, three crowdsourced judges were asked to whether to select one of Strongly Agree, Agree, Neutral, Disagree, Strongly Disagree in relation to the following statements: 1) The response is \emph{relevant} and appropriate. 2)The response is \emph{interesting} and informative. The final results were converted into a numerical score of 1 (Strongly Disagree) through 5 (Strongly Agree), and $95\%$ confidence intervals were computed using 10,000 iterations of bootstrap sampling. Because official organizer can only evaluate one system for each participant, we have only our \emph{Retrieval} system evaluated. From the results of Table \ref{tab:subject}, we can find that \emph{Seq2Seq} baseline system outperformed our \emph{Retrieval} system on all three metrics, which is not consistent with the objective comparison results in Table \ref{tab:object}
and Table \ref{tab:object_diversity}. The mismatching results between objective and subjective results indicate the difficulty of conversation response evaluation.
According to the official evaluation results, our \emph{Retrieval} system ranked third among all participant systems, which is not equivalent to its performance on objective evaluations. We think it may be attribute to that our \emph{Retrieval} system is not as flexible as generative models and it does not utilize the textual facts, which can be confirmed in some way by the objective evaluation results in Table \ref{tab:object}, i.e, the generative \emph{VariGen} system performed better than \emph{Retrieval} system on both NIST and BLEU metrics.

\section{Conclusion}
In this paper, we present our conversation response generation systems for DSTC7-Track 2. According to the official evaluation results, our proposed systems obtained best diversity performance on Entropy metrics among all participant systems, and at the same time, our systems also ranked second on NIST metrics, which shows that our proposed systems are promising to be developed further for conversation response generation.

\section{ Acknowledgments}
This work was partially funded by the National Nature Science Foundation of China (Grant No. U1636201).

\bibliography{dstc7}
\bibliographystyle{aaai}

\end{document}